\documentclass[conference]{IEEEtran}

\usepackage{graphicx} 
\usepackage{caption}
\usepackage{amsmath}
\usepackage{listings}
\usepackage{multirow}
\usepackage{cite}
\usepackage{caption}
\usepackage{subcaption}
\usepackage{tcolorbox}
\usepackage{hyperref}
\usepackage{url}
\usepackage{listings}
\usepackage{booktabs}
\usepackage{color}
\usepackage{soul}

\title{AST-Enhanced or AST-Overloaded? The Surprising Impact of Hybrid Graph Representations on Code Clone Detection}

\author{
    Zixian Zhang\textsuperscript{1,2}, Takfarinas Saber\textsuperscript{1,3} \\
    \textsuperscript{1}University of Galway, Ireland \\
    \textsuperscript{2}CRT-AI, Irish National Centre for Research Training in Artificial Intelligence \\
    \textsuperscript{3}Lero, the Research Ireland Centre for Software\\
    Email: z.zhang15@universityofgalway.ie, takfarinas.saber@universityofgalway.ie
}

\begin{document}

\maketitle

\begin{abstract}


As one of the most detrimental code smells, code clones significantly increase software maintenance costs and heighten vulnerability risks, making their detection a critical challenge in software engineering. Abstract Syntax Trees (ASTs) dominate deep learning-based code clone detection due to their precise syntactic structure representation, but they inherently lack semantic depth. Recent studies address this by enriching AST-based representations with semantic graphs, such as Control Flow Graphs (CFGs) and Data Flow Graphs (DFGs). However, the effectiveness of various enriched AST-based representations and their compatibility with different graph-based machine learning techniques remains an open question, warranting further investigation to unlock their full potential in addressing the complexities of code clone detection. In this paper, we present a comprehensive empirical study to rigorously evaluate the effectiveness of AST-based hybrid graph representations in Graph Neural Network (GNN)-based code clone detection. We systematically compare various hybrid representations ((CFG, DFG, Flow-Augmented ASTs (FA-AST)) across multiple GNN architectures. Our experiments reveal that hybrid representations impact GNNs differently: while AST+CFG+DFG consistently enhances accuracy for convolution- and attention-based models (Graph Convolutional Networks (GCN), Graph Attention Networks (GAT)), FA-AST frequently introduces structural complexity that harms performance. Notably, GMN outperforms others even with standard AST representations, highlighting its superior cross-code similarity detection and reducing the need for enriched structures.

\end{abstract}


\section{Introduction}
\label{sec:introduction}

Code clone detection is a fundamental task in software engineering, aimed at identifying duplicated or highly similar code fragments within a software repository~\cite{saini2018code}. Code clones can arise due to several development practices such as copy-pasting, reusing code templates, or implementing similar functionalities across different projects. While code duplication can improve development speed in the short term, it often leads to maintainability issues, increased technical debt, and security vulnerabilities~\cite{fowler2018refactoring}. Detecting and managing code clones is crucial for ensuring software quality, facilitating refactoring, and preventing unintended inconsistencies that may introduce bugs~\cite{saini2018code}.

A key aspect of code clone detection is the representation of source code. Various code representations have been proposed to capture the syntactic and semantic features of programs, enabling more effective analysis. Among these representations, the Abstract Syntax Tree (AST) is one of the most widely used due to its ability to capture the syntactic structure of programs while being easy to extract~\cite{kaur2023systematic}. ASTs abstract away surface-level variations, allowing models to focus on structural similarities rather than specific token sequences. However, despite its advantages, research has shown that AST-based representations primarily encode syntactic information and often fail to capture deeper semantic relationships in code~\cite{liu2023learning}.

To address these limitations, many studies have attempted to enhance AST-based graph structures by incorporating additional control and data flow information, leveraging Graph Neural Networks (GNNs) for code clone detection. A pioneering study in this direction was conducted by Wang et al. \cite{wang2020detecting}, who introduced a flow-augmentation technique that integrates ASTs with control and data flow information, thereby improving the detection of semantic code clones. Subsequent research has built upon this idea, enriching AST representations with handcrafted control and data flow information to combine both syntactic and semantic aspects of code \cite{fang2020functional, lu2021code, zhao2022precise, xu2021sccd, liu2023learning}. With the advancement of cross-language code clone detection, this approach has also been extended to that domain \cite{mehrotra2023improving, swilam2023cross}.

Despite extensive research in this field, the impact of augmenting AST-based representations with control and data flow information has not been systematically examined. In particular, ablation studies assessing the contribution of control and data flow integration within AST structures remain largely unexplored. Furthermore, the additional computational overhead introduced by incorporating control and data flow information—an essential consideration in the development of real-world applications—has received limited attention in existing research.

In this study, we conduct an empirical analysis to evaluate the effectiveness of various AST-based hybrid graph representations in GNN-based code clone detection. We provide a detailed investigation into how different edge representations in AST-based graphs impact both detection accuracy and computational efficiency, offering valuable insights to the open-source research community. Specifically, our research aims to answer the following questions:

\begin{itemize}
    \item \textbf{RQ1: Which AST-based hybrid graph representation and GNN architecture combination is most effective for code clone detection?}
    This research question aims to evaluate the impact of different AST-based hybrid graph structures (e.g., AST + CFG, AST + DFG, AST + FA) on code clone detection performance. We systematically compare these representations across multiple GNN architectures, including GCN, GAT, GGNN, and GMN. The analysis focuses on assessing their effectiveness in terms of accuracy, recall, and precision for detecting different types of code clones.
    \item \textbf{RQ2: What is the computational overhead of different AST-based hybrid representations?}
    This question investigates the trade-offs between detection performance and computational cost when incorporating additional structural information into AST-based hybrid graphs. We analyze key efficiency metrics such as memory consumption, graph density, generation time, and inference time to assess the feasibility of employing enriched representations in real-world applications.
\end{itemize}

In summary, our main contributions are as follows:
\begin{itemize}
    \item We conduct a systematic evaluation of various AST-based hybrid graph structures and their effectiveness across multiple GNN architectures, including GCN, GAT, GGNN, and GMN, for code clone detection. Our analysis provides valuable insights into the impact of different hybrid representations on detection accuracy and the comparative performance of different GNN models.
    
    \item We analyze the computational overhead associated with different AST-based hybrid representations, examining factors such as memory consumption, graph density, generation time, and inference time. This evaluation provides practical insights into the trade-offs between detection performance and computational efficiency in real-world applications.

    \item We present an open-source resource encompassing dataset allocation, graph construction methodologies, hybrid graph combinations, and model implementations. This resource facilitates further research by enabling the exploration of diverse hybrid graph representations and the development of more efficient GNN-based approaches for code clone detection. The resource is publicly available at \url{https://github.com/ZixianReid/semantic_graph_code_code_clone}.
\end{itemize}
This paper is organized as follows: Section~\ref{sec: background} introduces the necessary background concepts. Section~\ref{sec:related_work} provides an overview of related work. Section~\ref{sec:Methodology} details our experimental setup. Section~\ref{sec:Evaluation Analysis} presents the experimental results. Section~\ref{sec:Threats to Validity} discusses potential threats to the validity of our study. Section~\ref{sec:discussion} offers a discussion of key insights and implications. Section~\ref{sec:conclusion} concludes this~study.

\section{Background}
\label{sec: background}
In this section, we present the background necessary for the understanding of this work in three parts: Code Clone Detection, Source Code Representations, and Graph Neural Networks.

\subsection{Code Clone Detection}
Code clone detection aims to identify similar or duplicate code fragments. Code clones are typically categorized into four types \cite{choi2023investigating}:
\begin{itemize}
    \item \textbf{Type-1:} Code fragments that are identical except for superficial differences such as formatting and comments.
    \item \textbf{Type-2:} Code fragments that exhibit minor modifications, such as renamed variables, altered data types, or changes in literals, while maintaining structural similarity.
    \item \textbf{Type-3:} Code fragments that have undergone significant structural modifications, including statement insertions, deletions, or reordering, yet still preserve core functionality.
    \item \textbf{Type-4:} Code fragments that achieve the same functionality through different implementations.
\end{itemize}
This study primarily focuses on Type-4 code clones, which are more challenging to detect due to their structural differences

\subsection{Source Code Representations}
In this study, we explore four distinct source code representations that have been widely utilized for code clone detection: Abstract Syntax Tree (AST), Control Flow Graph (CFG), Flow-Augmented Abstract Syntax Tree (FA-AST), and Data Flow Graph (DFG).

\begin{itemize}
    \item \textbf{Abstract Syntax Tree (AST):} AST is the syntactical graph-based representation of code fragments, which is one of the most popular representations in code analysis. They abstract away surface-level variations, allowing models to focus on structural similarities rather than specific token sequences. 
    \item \textbf{Control Flow Graph (CFG):} CFGs capture the execution flow of a program, representing how different statements and expressions interact based on control structures. CFGs are particularly useful for identifying logical similarities between code fragments.
    \item \textbf{Data Flow Graph (DFG):} DFGs model the dependencies between variables and expressions by tracking how data propagates through the program. They are beneficial in detecting clones with similar computational logic but different syntactic structures.
    \item \textbf{Flow-Augmented Abstract Syntax Tree (FA-AST):} Wang et al. \cite{wang2020detecting} constructed FA-AST by augmenting AST with explicit control and data flow edges to better capture semantic information. While this modification improves the AST's ability to convey program behavior, it also introduces increased computational and structural complexity. For consistency with our hybridization notation, we refer to this representation in our experiments as FA.
\end{itemize}
In this study, we fuse these code representations in different combinations to evaluate their impact on code clone detection. By constructing various hybrid representations (e.g., AST~+~CFG, AST~+~DFG, and AST~+~FA~+~CFG~+~DFG), we aim to analyze the role of each representation in capturing syntactic and semantic similarities while also investigating the trade-offs between detection accuracy and computational complexity.

\subsection{Graph Neural Networks (GNNs)}
Unlike Convolutional Neural Networks (CNNs) or Recurrent Neural Networks (RNNs), GNNs exploit the underlying graph topology to learn meaningful node or graph-level representations. A key principle of most GNN architectures is the message-passing paradigm, where node embeddings are iteratively updated by aggregating and transforming information from their neighbors. All GNNs used in this study follow this message-passing framework.
\begin{itemize}
    \item \textbf{Graph Convolutional Network (GCN):} GCN \cite{kipf2017semi} is a fundamental graph-based model that applies convolutional operations to aggregate information from neighboring nodes, enabling efficient representation learning on graph-structured data. It captures structural dependencies within a graph by iteratively updating node representations based on their local neighborhoods.
    \item \textbf{Graph Attention Network (GAT):} GAT \cite{velivckovic2018graph} enhances graph representation learning by introducing attention mechanisms that assign different importance weights to neighboring nodes during message passing. This adaptive weighting allows the model to focus on the most relevant structural elements
    \item \textbf{Graph Gated Neural Network (GGNN):} GGNN \cite{li2015gated} extends traditional GNNs by incorporating gated recurrent units (GRUs) to model long-range dependencies in graph structures. This approach enables better information propagation across large and complex graphs, making it useful for applications that require deeper contextual understanding
        \item \textbf{Graph Matching Network (GMN):} GMN \cite{li2016gated} is designed for comparing graph structures by learning a similarity function between node representations. It is particularly effective in tasks that require assessing relational patterns, such as similarity learning and structural alignment, where graph-level relationships must be preserved.
\end{itemize}

\section{Related work}
\label{sec:related_work}

In this section, we present the work related to our study of using AST-based hybrid graph representations with GNNs and other works leveraging alternative representations and machine learning techniques for code clone detection.

\textbf{AST-based Hybrid Graph Representations} are widely utilized in code clone detection as they integrate multiple code representations to provide a comprehensive analysis of both the syntactic and semantic aspects of code fragments. One of the pioneering works in this domain is FA-AST by Wang et al. \cite{wang2020detecting}, which employs a Graph Neural Network (GNN) on flow-augmented abstract syntax trees to effectively capture syntactic and semantic similarities in code fragments. In FA-AST, flow edges such as \texttt{nextUse}, \texttt{nextToken}, and \texttt{nextSibling} are directly embedded into the AST to enrich it with additional semantic signals. However, their representation is tightly coupled and fixed within the AST structure.

Beyond FA-AST, a common approach to enhancing AST-based representations involves the integration of the AST, CFG, and DFG. This hybrid strategy facilitates a more comprehensive analysis of program functionalities by capturing syntactical structures through the AST, control dependencies via the CFG, and data dependencies using the DFG. Various studies have employed such hybrid representations in both single-language and cross-language code clone detection, leveraging different GNN architectures. For instance, Zhao et al. \cite{zhao2022precise} integrate the AST and CFG with hierarchical dependencies, employing GAT to improve detection accuracy. Similarly, Fang et al. \cite{fang2020functional} and Xu et al. \cite{xu2021sccd} have adopted this fusion methodology in their studies.

Some research has explored more comprehensive combinations of AST, CFG, and DFG. For example, Yuan et al. \cite{yuan2022java} introduce an intermediate code-based graph representation that integrates these three components, thereby enhancing the identification of functional code clones. Additionally, Liu et al. \cite{liu2023learning} propose TAILOR, which incorporates AST, CFG, and DFG to improve the detection of functionally similar code fragments. Similar hybrid graph representations have also been extended to cross-language code clone detection, as demonstrated by Mehrotra et al. \cite{mehrotra2023improving} and Swilam et al. \cite{swilam2023cross}.

While these studies highlight the significance of hybrid graph representations as an innovation in clone detection, their specific impact remains insufficiently explored. A major challenge in this research area is the variation in experimental settings across different studies, including the use of distinct GNN architectures, datasets, and embedding techniques. Consequently, the direct contribution of hybrid representations to code clone detection remains obscured despite their widespread adoption. Moreover, when evaluating the effectiveness of code representations, computational overhead is critical in constructing efficient code clone detection tools. However, within the scope of existing research, only a limited number of studies have assessed the computational performance of AST-based approaches from a comprehensive perspective \cite{liu2023learning, fang2020functional}. Prior work has primarily focused on inference and training time, often comparing proposed methods against other code clone detection tools. However, key aspects in AST-based hybrid representations, such as extra computational overhead and storage requirements, have been largely overlooked despite their significance in real-world applications. Addressing these gaps is crucial for advancing more efficient and scalable code clone detection methods.

\begin{figure*}[!htbp]
    \centerline{\includegraphics[width = 1.\linewidth]{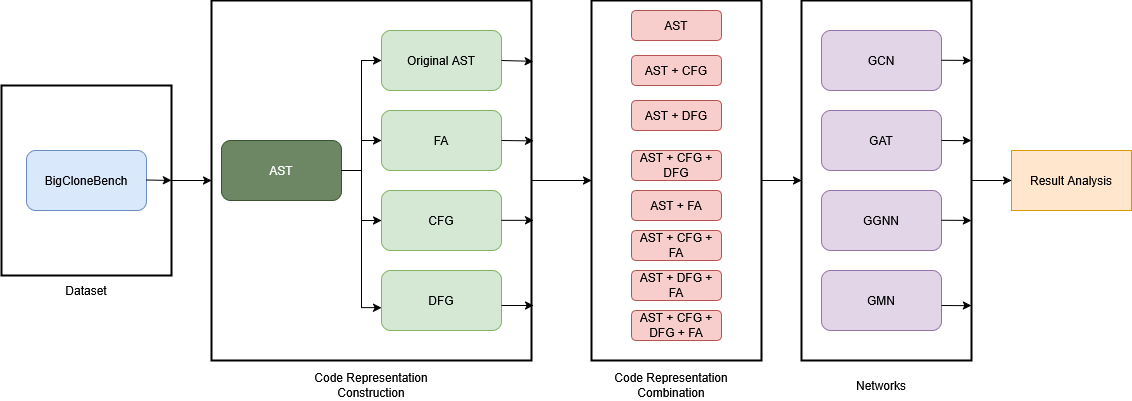}}
    \caption{Methodology Employed in Our Study.}
    \label{fig:The_process_of_methodology}
\end{figure*}

\textbf{Evaluation of Code Representation in Code Clone Detection} is crucial for understanding the effectiveness of different code structures in capturing similarities. Most empirical studies in this domain have been tool-based, focusing on evaluating the performance of specific clone detection tools \cite{svajlenko2015evaluating, svajlenko2016bigcloneeval, choi2023investigating, van2020clone, choi2023investigating}. Recent research has explored the impact of different code representations on clone detection performance. Wang et al. \cite{wang2023comparison} reproduce 12 clone detection algorithms categorized into text-based, token-based, tree-based, and graph-based approaches, revealing that token- and AST-based methods excel at detecting simple clones but struggle with semantic ones. In contrast, CFG- and PDG-based approaches achieve higher recall for complex clones but incur greater computational costs. Zubkov et al. \cite{zubkov2022evaluation} evaluated contrastive learning methods for code clone detection, comparing SimCLR, SwAV, and Moco across text, AST, and graph-based representations. Their findings show that graph-based models outperform others, with SimCLR and SwAV achieving the best results, while Moco demonstrates robustness. 
However, no studies focus on the impact of AST-based representations on code clone detection. Moreover, while some of these studies analyze the performance of different code representations, they primarily focus on reproducing existing algorithms where code representations are applied. However, other factors, such as network architecture, training strategies, and hyperparameter selection, introduce biases that are often overlooked. To address these gaps, this study systematically investigates the impact of different AST-based representations across various network architectures. By implementing diverse AST-based representations under identical experimental conditions, we aim to isolate the effects of code representation itself, ensuring a fair evaluation of its influence on code clone detection performance.

\section{Methodology}
\label{sec:Methodology}

This section details the evaluation process used in this study. A visual representation of the methodology is provided in Figure~\ref{fig:The_process_of_methodology}.

\subsection{Problem Formulation}
The code clone detection problem is defined as follows: Given two code fragments, $C_i$ and $C_j$, the goal is to associate them with a label $y_{ij}$ that determines whether they are clones:
\begin{itemize}
    \item $y_{ij} = 1$ indicates that $C_i$ and $C_j$ are clones,
    \item $y_{ij} = 0$ indicates that they are not clones.
\end{itemize}

Let $s_{ij} \in [0, 1]$ represent a similarity score between the given pair of code fragments:

\begin{itemize}
    \item $s_{ij} = 1$ indicates perfect similarity (i.e., an exact clone),
    \item $s_{ij} = 0$ indicates complete dissimilarity (i.e., a non-clone).
\end{itemize}

To assess classification performance, we binarized the predicted similarity score $s_{ij}$ using a fixed \textit{classification threshold} $\sigma$, following standard practices in the literature. Formally:
\begin{equation}
y_{ij} =
\begin{cases} 
1, & \text{if } s_{ij} > \sigma \text{ (clone pair)} \\
0, & \text{otherwise (non-clone pair)}
\end{cases}
\end{equation}

This choice aligns with prior work such as Wang et al. \cite{wang2020detecting} and Zhang et al.~\cite{zhang2023efficient}.

\subsection{Dataset Selection and Filtering}

We use BigCloneBench (BCB) \cite{svajlenko2014towards}, one of the largest and most widely used benchmarks for code clone detection. Specifically, we adopt the balanced and filtered version introduced by Wei and Li \cite{wei2017supervised}, which excludes unlabeled code fragments (i.e., those not explicitly tagged as true or false clone pairs). However, this version lacks clone type labels and similarity scores. To recover this information, we merge it with the original BCB by aligning code fragment IDs, thereby retaining the benefits of a balanced dataset while restoring rich annotation metadata (e.g., clone type and similarity scores). Table~\ref{tab:demo_bcb} summarizes the dataset statistics: the 8,876 code fragments were paired to form labeled positive and negative clone pairs used across training, validation, and test sets.


\begin{table*}[!htbp]
\centering
\caption{Clone Type Demographics in~BigCloneBench.}
\label{tab:demo_bcb}
\begin{tabular}{|l|l|l|l|l|}
\hline
\textbf{Dataset}    & \textbf{Code fragments}        & \textbf{Average Number of Line of Codes}  & \textbf{Positive Pairs} & \textbf{Negative Pairs} \\ \hline
Training   & \multirow{3}{*}{8876} & \multirow{3}{*}{34.17} & 610247         & 601376         \\ \cline{1-1} \cline{4-5} 
Testing    &                       &                        & 73796          & 348982         \\ \cline{1-1} \cline{4-5} 
Validation &                       &                        & 54818          & 348298         \\ \hline
\end{tabular}
\end{table*}

\subsection{Source code Representations}

Abstract Syntax Trees (ASTs) serve as the fundamental representation in this study. All additional representations, including Control Flow Graphs (CFG), Data Flow Graphs (DFG), and Flow-Augmented ASTs (AST + FA), are constructed based on ASTs. To extract ASTs from Java programs, we utilize the Python package Javalang. AST + FA representations are constructed following the methodology outlined in the original FA-AST paper by Wang et al. \cite{wang2020detecting}.

For CFG and DFG construction, we adopt a methodology similar to prior studies that leverage AST-based hybrid representations~\cite{liu2023learning, mehrotra2023improving}. The construction process consists of two main steps:

\begin{itemize}
 \item \textbf{AST Construction:} We first extract the AST structure using the \texttt{Javalang} library, including node and edge information.
    \item \textbf{Dependency Identification and Graph Augmentation:} Based on the AST, we identify two types of dependencies:
    \begin{itemize}
        \item \textit{Control Dependencies:} We traverse the AST to locate control structures (e.g., \texttt{if}, \texttt{while}, \texttt{for}, \texttt{return}) and generate control flow edges that connect nodes based on the execution order of statements.
        \item \textit{Data Dependencies:} We identify variable definitions and subsequent uses within the same function. A backward-tracking algorithm is applied from each variable use to find the nearest dominating definition along the AST traversal path. If such a definition is found, we add a directed edge from the defining node to the usage node, forming the DFG.
    \end{itemize}
    The resulting control and data flow relationships are incorporated into the original AST structure as additional directed edges with distinct edge types.
\end{itemize}

\begin{figure}[!htbp]
    \centerline{\includegraphics[width = 0.9\linewidth]{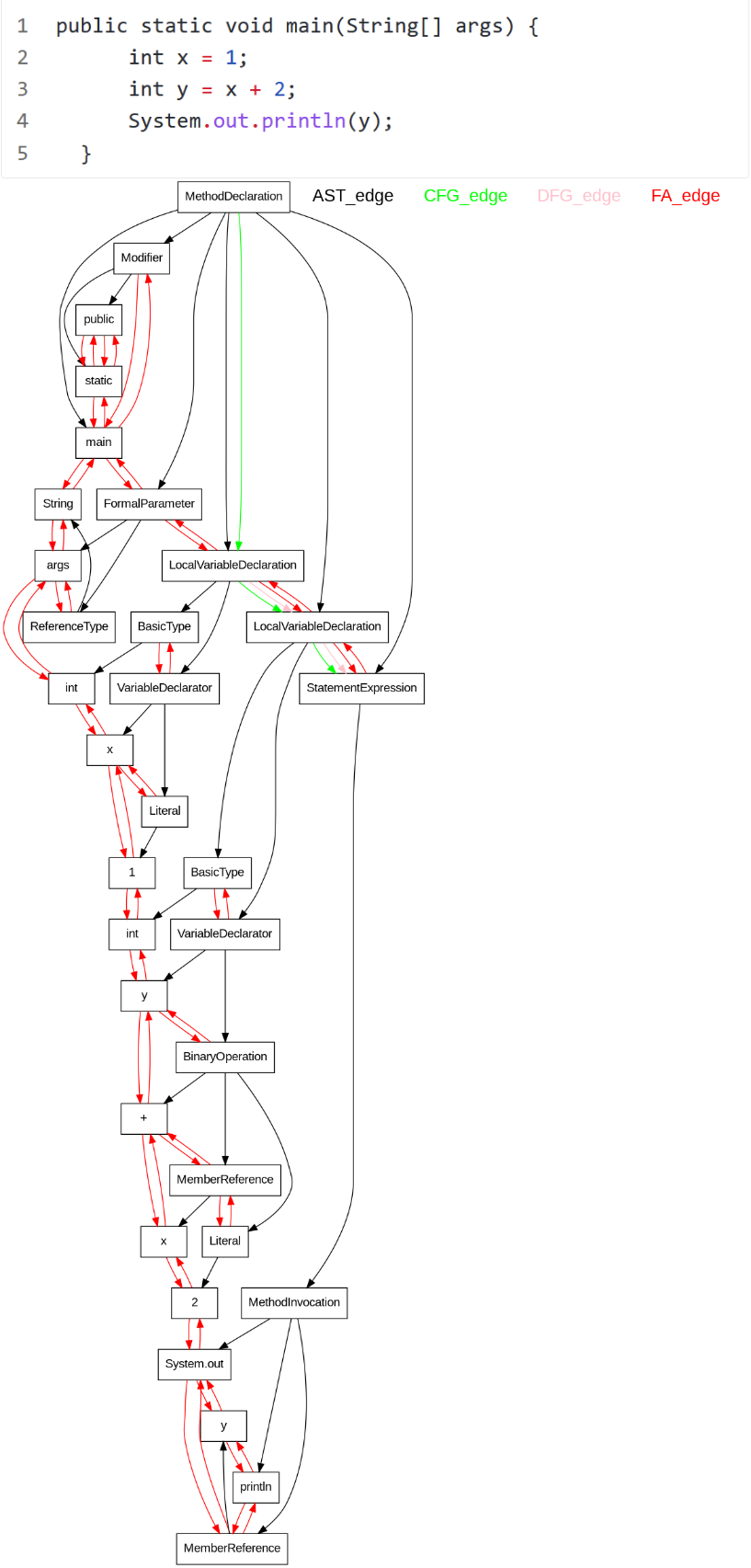}}
    \caption{AST-based Hybrid Code Representation of the sample function.}
    \label{fig:The process of methodology in this study}
\end{figure}

Figure~\ref{fig:The process of methodology in this study} presents a visual representation of the constructed model for a sample code fragment.

\begin{figure*}[!htbp]
\centering
    \begin{tabular}{c}
         \includegraphics[width = 0.9\linewidth]{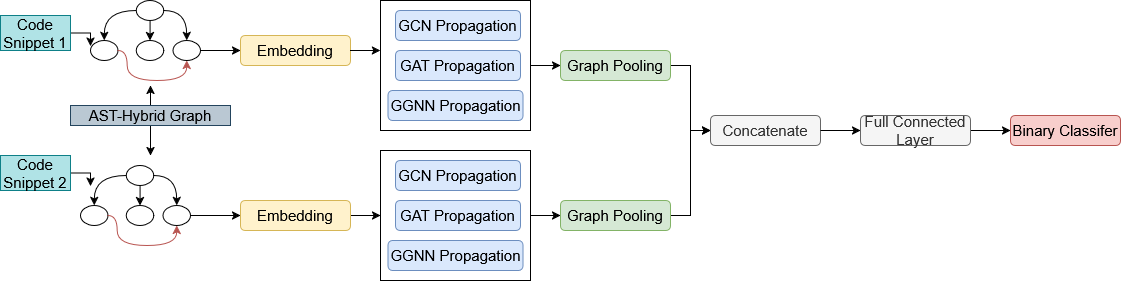}
         \\
         a) Using GCN, GAT, or GGNN
         \\
         \\
        \includegraphics[width = 0.9\linewidth]{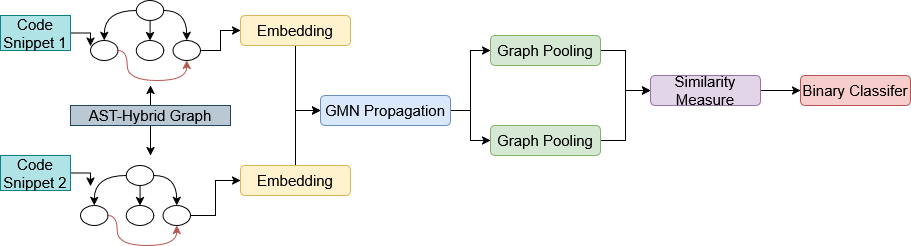}
        \\
        b) Using GMT
         \\
    \end{tabular}
    \caption{Code Clone Detection Pipeline with the Considered GNNs: GCN, GAT, and GGNN (Top Figure); GMN (Bottom Figure).}
    
    \label{fig:Illustration_gnns}
\end{figure*}

\subsection{Network Configuration}
We implement all GNNs with PyTorch \cite{paszke2019pytorch} and its extension library Pytorch Geometric \cite{fey2019fast}. For a fair comparison of various GNNs, we maintain a consistent architectural configuration. The overall architecture consists of an embedding layer, a graph propagation layer, a global pooling mechanism, and a classification head.

\begin{enumerate}
    \item \textbf{Input Representation and Embedding:} We employ an embedding layer to transform node representations into a continuous vector space. Specifically, we employ the embedding function from PyTorch \cite{paszke2019pytorch}, which implements a lookup table that stores embeddings for a fixed vocabulary size and dimensionality.  Similarly, edge attributes are also embedded using a separate embedding layer, where each edge type is mapped to a corresponding feature vector. 
    \item \textbf{Graph Propagation Layers:} We utilize three standard GNN propagation layers from the PyTorch Geometric library to implement propagation layer of GCN, GAT and GGNN, whereas we implement the propagation layer of GMN in the work of Wang et al.~\cite{wang2020detecting}.
    \item \textbf{Graph Pooling:} We apply global attention pooling to obtain a fixed-size graph representation.
    \item \textbf{Classification Head:} For GCN, GAT, and GGNN, the final node embeddings of the two graphs are concatenated before being processed through a fully connected feed-forward network. This approach merges the graph representations into a shared feature space before computing a similarity score. For GMN, the pooled graph representations are directly compared to determine their similarity.
\end{enumerate}
The diagrammatic sketch of GNNs architectures is shown in Figure~\ref{fig:Illustration_gnns}.

\subsection{Evaluation Metrics}

We use multiple evaluation metrics to assess the effectiveness of different AST-based hybrid graph representations in code clone detection. These metrics include Precision, Recall, and F1-score.








To complement accuracy-based evaluation, we report four additional metrics to assess the efficiency and structural properties of each graph representation: \textbf{Generation Cost}, \textbf{Storage Cost}, \textbf{Average Graph Density}, and \textbf{Inference Time}.
\textbf{Generation Cost} refers to the time required to construct the graph representations from raw code in the BigCloneBench dataset, which consists of $8,876$ code fragments.

\textbf{Storage Cost} represents the total memory footprint of the generated graphs for all code fragments in BigCloneBench.

\textbf{Average Graph Density} quantifies the overall connectivity of graphs in the dataset by averaging the density of individual graphs:


\begin{equation}
\text{Average Graph Density} = \frac{1}{N} \sum_{i=1}^{N} \frac{|E_i|}{|V_i| (|V_i| - 1)}
\end{equation}

where:
\begin{itemize}
    \item \( N \) is the total number of graphs in the dataset,
    \item \( |E_i| \) is the number of edges in the \( i \)-th graph,
    \item \( |V_i| \) is the number of nodes in the \( i \)-th graph.
\end{itemize}

A higher average graph density suggests a more interconnected structure, which may enhance representational power but also increase computational overhead.

\textbf{Inference Time} evaluates the efficiency of utilizing these graph representations for code clone detection. In this study, we measure the total time required for GMN to process the test set, which contains 422,780 code fragment pairs.

\subsection{Experimental Settings}
We implement the models using PyTorch \cite{paszke2019pytorch} and PyTorch Geometric \cite{fey2019fast}. All experiments are conducted on a machine equipped with an Intel i9-13900K, 32GB RAM, and an NVIDIA RTX A4000 WITH 16GB memory. The models are trained using the Adam optimizer with an initial learning rate of 0.0005 and a weight decay of $1\times10^{-4}$. To dynamically adjust the learning rate, we apply a learning rate scheduler that reduces the learning rate by a factor of 0.5 if the validation loss does not improve for two consecutive epochs, with a lower bound of $1\times10^{-6}$. We train the model for 20 epochs with a batch size of 32, optimizing using Mean Squared Error (MSE) Loss. Each input consists of two code fragments represented as AST-based hybrid graphs, which are embedded into a 100-dimensional space before being processed by a 4-layer GNN. The model supports multiple architectures, including GCN, GAT, GGNN, and GMN, with each variant propagating node embeddings differently. The dataset is divided into training (80\%), validation (10\%), and testing (10\%) sets.

\section{Experimental Results and Analysis}
\label{sec:Evaluation Analysis}
In this section, we aim to answer the two research questions defined above by describing and analyzing the results of our experimental study.

\begin{table*}[!htbp]
    \centering
    \caption{Performance of Different AST-Based Hybrid Graph Representations Across GNNs.}
    \begin{subtable}[!htbp]{0.45\textwidth}
        \centering
        \caption{Graph Matching Network (GMN)}
        \begin{tabular}{|l|c|c|c|}
            \hline
            \textbf{Representation} & \textbf{Precision} & \textbf{Recall} & \textbf{F1-score} \\
            \hline
            AST & \textbf{0.986} & 0.922 &  0.953 \\
            AST + CFG & 0.977 & 0.910 & 0.943 \\
            AST + DFG & 0.967 & 0.934 & 0.950 \\
            AST + CFG + DFG & 0.970 &  \textbf{0.928} & 0.948 \\
            AST + FA & 0.982 & 0.920 & 0.950 \\
            AST + FA + CFG & \textbf{0.986} & 0.925 & \textbf{0.954} \\
            AST + FA + DFG & 0.982 & 0.908 & 0.943 \\
            AST + FA + CFG + DFG& 0.975 & 0.926 & 0.950 \\
            \hline
        \end{tabular}
    \end{subtable}
    \hfill
    \begin{subtable}[!htbp]{0.45\textwidth}
        \centering
        \caption{Graph Convolutional Network (GCN)}
        \begin{tabular}{|l|c|c|c|}
            \hline
            \textbf{Representation} & \textbf{Precision} & \textbf{Recall} & \textbf{F1-score} \\
            \hline
            AST & 0.910 & 0.829 & 0.868 \\
            AST + CFG & \textbf{0.951} & 0.845 & 0.896 \\
            AST + DFG & 0.939 & 0.821 & 0.876 \\
            AST + CFG + DFG & 0.950 & \textbf{0.855} & \textbf{0.900} \\
            AST + FA & 0.920 & 0.842 & 0.879 \\
            AST + FA + CFG & 0.928 & 0.812 & 0.866 \\
            AST + FA + DFG & 0.921 & 0.843 & 0.880 \\
            AST + FA + CFG + DFG & 0.924 & 0.844 & 0.882 \\
            \hline
        \end{tabular}
    \end{subtable}
    
    \vspace{0.5cm}

    \begin{subtable}[!htbp]{0.45\textwidth}
        \centering
        \caption{Graph Attention Network (GAT)}
        \begin{tabular}{|l|c|c|c|}
            \hline
            \textbf{Representation} & \textbf{Precision} & \textbf{Recall} & \textbf{F1-score} \\
            \hline
            AST & 0.904 & 0.842 & 0.872 \\
            AST + CFG &  \textbf{0.924} & 0.866 & 0.894 \\
            AST + DFG & 0.912 & 0.859 & 0.885 \\
            AST + CFG + DFG & 0.922 & \textbf{0.881} & \textbf{0.901} \\
            AST + FA & 0.923 & 0.860 & 0.891 \\
            AST + FA + CFG & 0.914 & 0.844 & 0.878 \\
            AST + FA + DFG & 0.921 & 0.853 & 0.885 \\
            AST + FA + CFG + DFG & 0.895 & 0.859 & 0.876 \\
            \hline
        \end{tabular}
    \end{subtable}
    \hfill
    \begin{subtable}[!htbp]{0.45\textwidth}
        \centering
        \caption{Graph Gated Neural Network (GGNN)}
        \begin{tabular}{|l|c|c|c|}
            \hline
            \textbf{Representation} & \textbf{Precision} & \textbf{Recall} & \textbf{F1-score} \\
            \hline
            AST & 0.907 & \textbf{0.890} & 0.900 \\
            AST + CFG & \textbf{0.935} & 0.865 & 0.900 \\
            AST + DFG & 0.912 & 0.889 & 0.900 \\
            AST + CFG + DFG & 0.923 & 0.884 & \textbf{0.903} \\
            AST + FA & 0.910 & 0.847 & 0.877 \\
            AST + FA + CFG & 0.919 & 0.859 & 0.888 \\
            AST + FA + DFG & 0.879 & 0.856 & 0.867 \\
            AST + FA + CFG + DFG & 0.891 & 0.843 & 0.866 \\
            \hline
        \end{tabular}
    \end{subtable}
    \label{tab:performance}
\end{table*}

\begin{figure*}[!htbp]
    \centering
    \begin{tabular}{ccc}
        \includegraphics[width=0.32\textwidth]{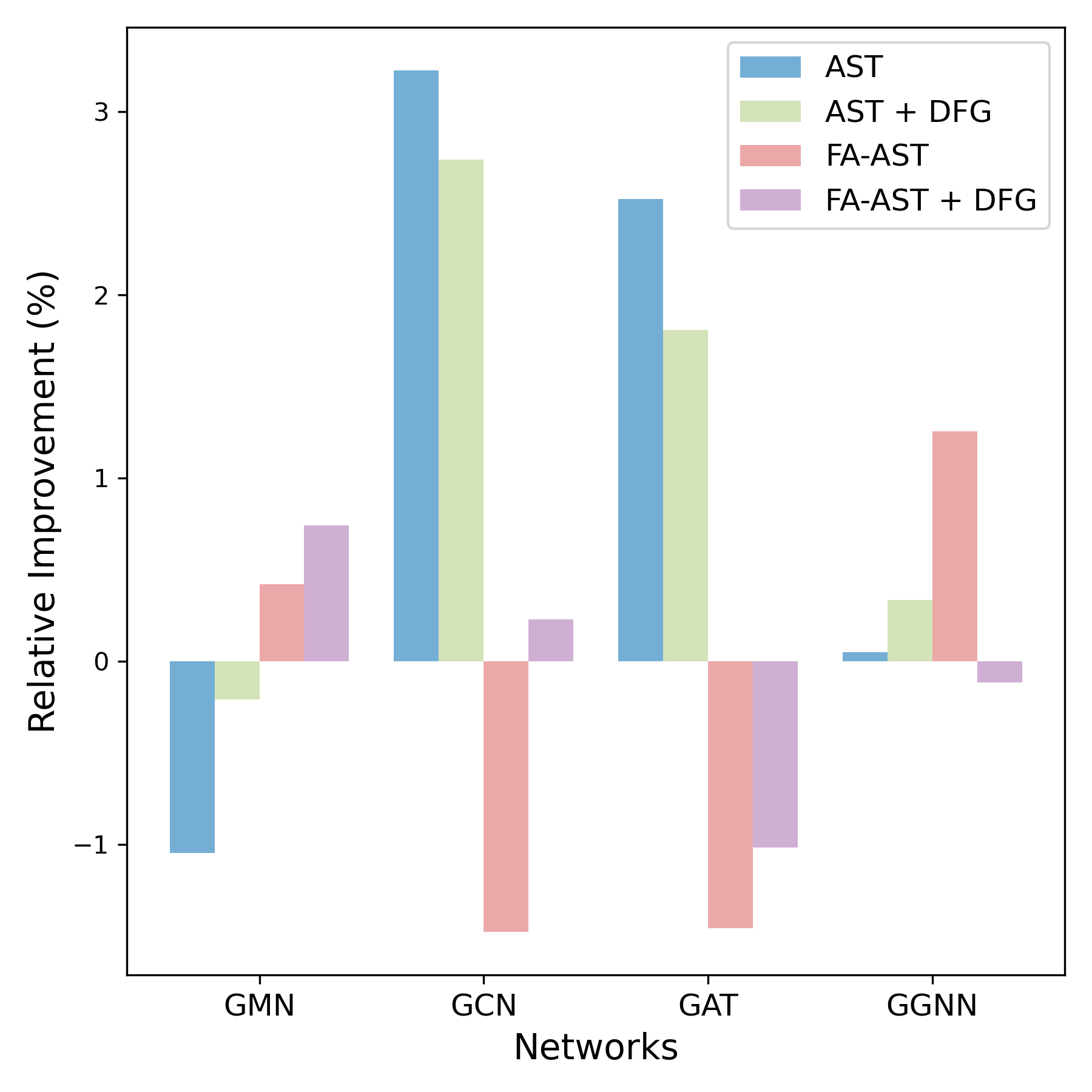} &
        \includegraphics[width=0.32\textwidth]{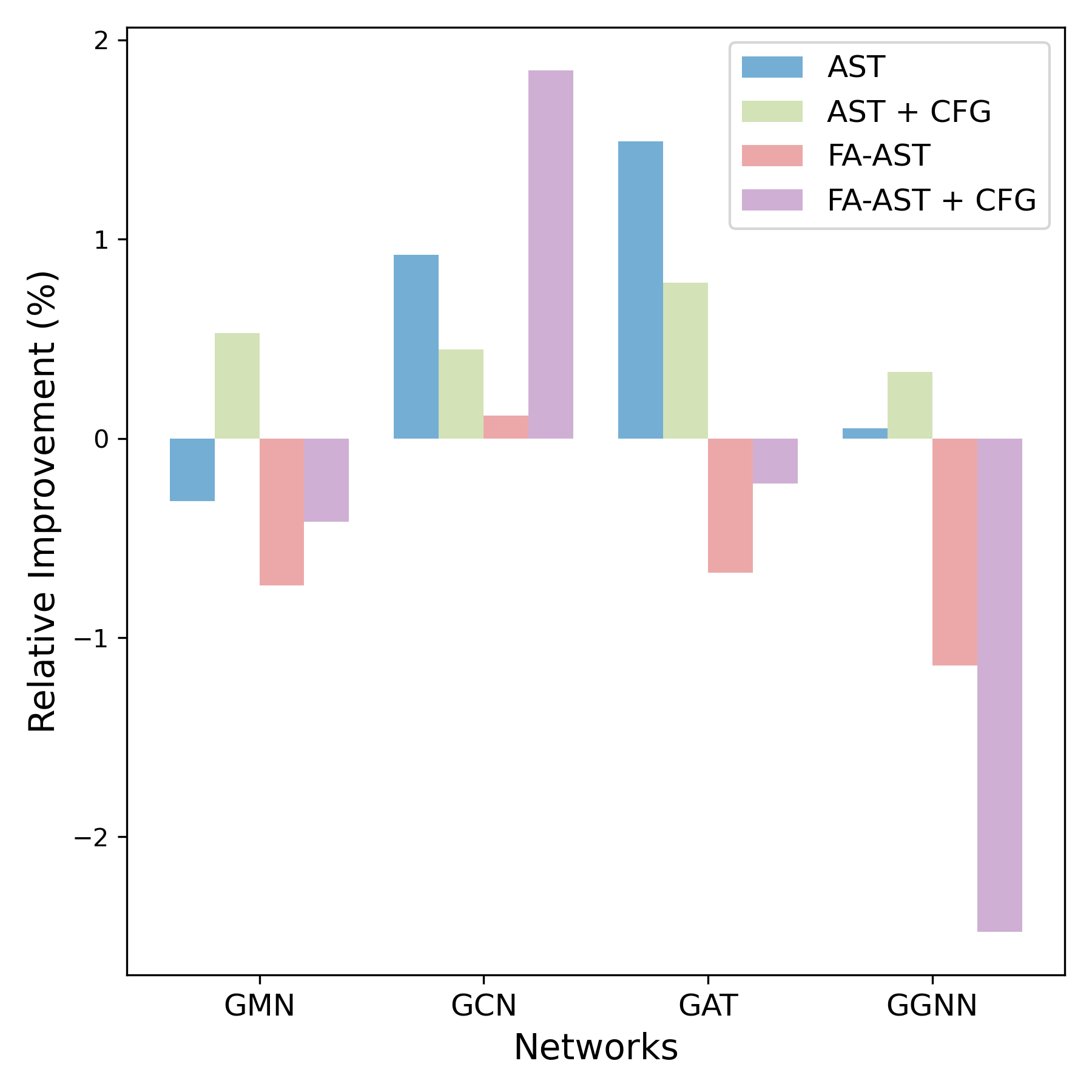} &
        \includegraphics[width=0.32\textwidth]{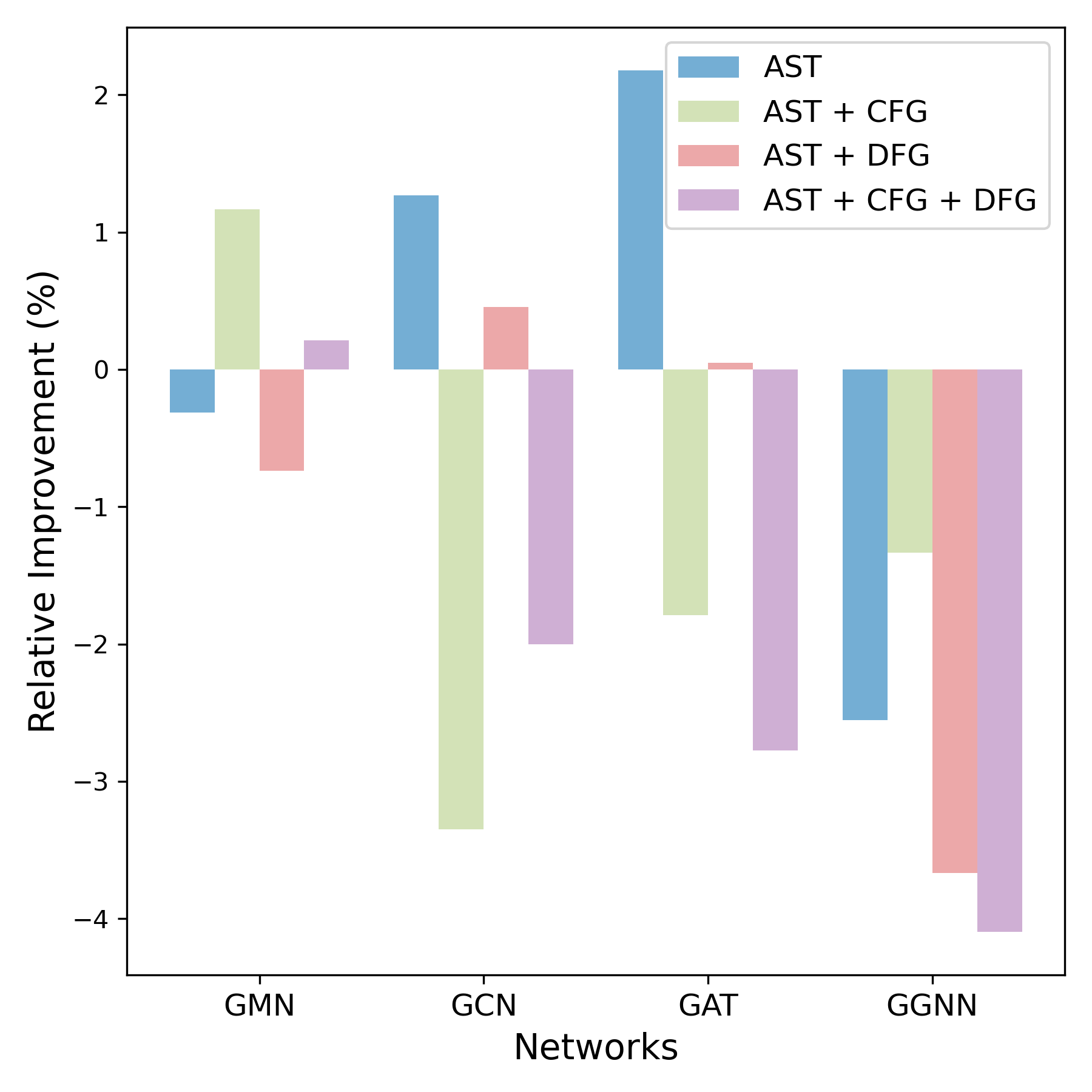} \\
        (a) Introducing CFG & (b) Introducing DFG & (c) Introducing FA \\
    \end{tabular}
    \caption{Relative F1-score Improvement Achieved By Alternatively Introducing the Representations CFG, DFG, or FA to Other Code Representations While Using the Different GNN Models.}
    \label{fig:relative_improvement}
\end{figure*}

\subsection{RQ1: Which AST-based hybrid graph representation and GNN architecture combination is most effective for code clone detection?}

We will attempt to answer this question in two phases: one which focuses on the most effective AST-based hybrid graph representation and another one that focuses on the most effective GNN model architecture.

\subsubsection{RQ1.1} \textit{Which AST-based hybrid graph representation is most effective in detecting code clones?}

To assess the effectiveness of different AST-based hybrid graph representations in detecting code clones, we conducted an empirical evaluation across four GNNs architectures: GMN, GCN, GAT and GMN. The performance of each hybrid representation is measured using Precision, Recall and F1-score, and the results are summarized in Table~\ref{tab:performance}.

We observe that different AST-based hybrid representations have varying impacts on different GNN architectures. Below, we analyze their effects on each model:
\begin{itemize}
    \item \textbf{GMN:} Our results indicate that while AST + FA + CFG achieves the highest F1-score and AST + CFG + DFG yields the highest recall, the standard AST representation attains the highest precision. Notably, the F1-score difference between AST and AST + FA + CFG is only 0.001, suggesting that the performance gain from hybrid representations is minimal. Furthermore, most AST-based hybrid representations result in a performance decline compared to the standard AST, highlighting that GMN is inherently effective at capturing syntactic patterns from ASTs without requiring additional flow-augmented structures. This finding suggests that while enriched AST structures may offer marginal benefits, GMN already excels in leveraging AST's syntactic features for code clone detection.
    \item \textbf{GCN:} Our results indicate that GCN achieves the highest F1-score and recall with AST + CFG + DFG, while AST + CFG attains the highest precision. Notably, most AST-based hybrid representations improve GCN's performance in code clone detection, demonstrating the benefits of incorporating additional semantics information. However, AST + FA + CFG leads to a decline in performance, suggesting that not all flow combinations are beneficial for convolutional feature propagation. These findings indicate that integrating semantic flow information, particularly through a combination of control and data flow graphs, enhances GCN's ability to capture meaningful structural dependencies.
    \item \textbf{GAT:} The results for GAT closely resemble those of GCN. We observe that GAT achieves the highest F1-score and recall with AST + CFG + DFG, while AST + CFG yields the highest precision. Notably, AST-based hybrid representations enhance GAT's performance in code clone detection to varying degrees, indicating that the integration of control and data flow information effectively supports attention-based learning. These findings suggest GAT effectively leverage the enriched structural dependencies provided by hybrid AST representations, leading to improved detection accuracy.
    \item \textbf{GGNN:} Our findings indicate that while AST + CFG + DFG achieves the highest F1-score, AST + CFG attains the highest precision, and the standard AST representation yields the highest recall. Additionally, the F1-scores among AST, AST + CFG, and AST + CFG + DFG are very close, suggesting that the performance differences between these representations are minimal. Unlike GCN and GAT, where FA-AST-based hybrids had varying effects, GGNN exhibits a noticeable decline in performance when using FA-AST-based hybrids, indicating that the introduction of flow-augmented AST structures may not be beneficial for recurrent architectures.
\end{itemize}

Moreover, beyond comparing the performance of different hybrid representations, we aim to assess the contribution of each individual representation to the overall performance improvement. To achieve this, we calculate the relative percentage improvement of each added representation.

The relative F1-score percentage improvement observed across GMN, GCN, GAT, and GGNN when integrating CFG, DFG and FA representations into various AST-based hybrid structures is illustrated in Figure~\ref{fig:relative_improvement}.

According to Figure~\ref{fig:relative_improvement}, it is evident that for GCN and GAT, the addition of all extra semantic information consistently leads to performance improvements. Specifically, CFG and DFG exhibit a similar impact, suggesting that both control flow and data flow information contribute positively to enhancing code representations for these models. Moreover, when both CFG and DFG are combined with AST, their synergistic effect further enhances performance, reinforcing the importance of incorporating both control and data flow structures into graph-based code analysis. However, when AST + FA is combined with either CFG or DFG, the performance generally deteriorates across all networks, indicating that FA edges may introduce noise or redundant connections that negatively impact learning.

A closer examination of AST + FA impact reveals that while it can enhance performance in specific cases (e.g., improving AST with GCN and AST + CFG with GMN), its overall contribution across different hybrid representations is largely negative. This effect is particularly pronounced with GGNN, where the addition of FA leads to a clear decline in performance across all representations. The negative impact of AST + FA suggests that the additional flow augmentation edges might interfere with the ability of message-passing GNNs to extract meaningful patterns, possibly due to an increase in graph complexity or redundant dependencies that hinder effective feature propagation.

\begin{table*}[!htbp]
    \centering
    \caption{Computational Overhead and Storage Cost of AST-Based Hybrid Graph Representations. Inference Time Measured using GMN.}
    \begin{tabular}{lcccc}
        \toprule
        \textbf{Code Representations} & \textbf{Generation Cost (s)} & \textbf{Graph Storage Cost (MB)} & \textbf{Average Graph Density} & \textbf{Inference Time (s)} \\
        \midrule
        AST                 & 14.055  & 527.54  & 0.0072  & 705.680  \\
        AST + CFG           & 14.224  & 591.00  & 0.0085  & 710.026  \\
        AST + DFG           & 305.095 & 533.38  & 0.0073  & 708.282  \\
        AST + CFG + DFG     & 305.407 & 596.85  & 0.0087  & 713.923  \\
        AST + FA              & 16.910  & 1152.00 & 0.0202  & 748.402  \\
        AST + FA + CFG        & 17.371  & 1215.44 & 0.0216  & 752.855       \\
        AST + FA + DFG        & 312.535 & 1157.83 & 0.0204  & 751.140       \\
        AST + FA + CFG + DFG  & 313.585 & 1221.40 & 0.0217  & 756.463      \\
        \bottomrule
    \end{tabular}
    \label{tab:graph_performance_Computational Overhead}
\end{table*}

\subsubsection{RQ1.2} \textit{Which GNN model architecture is more effective for code clone detection using AST-based hybrid graphs?}

\begin{figure}[!htbp]
    \centering
    \includegraphics[width=\linewidth]{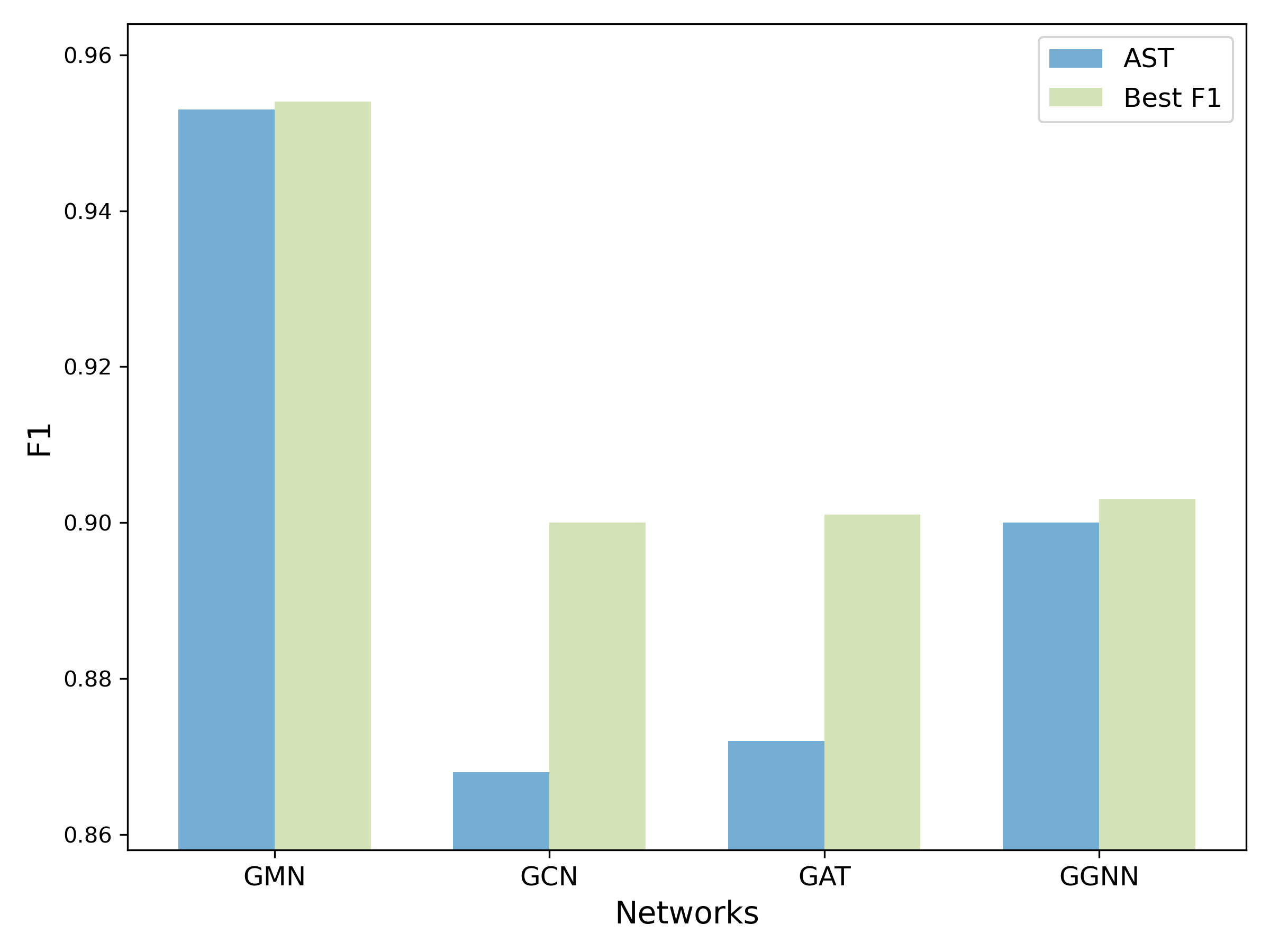}
    \caption{Performance Comparison Across Various GNNs.}
    \label{fig:network_comparsion}
\end{figure}

To assess the effectiveness of different GNN architectures in code clone detection, we compared the F1-scores of the standard AST representation and the best-performing AST-based hybrid representation for each network. The results are visualized in Figure~\ref{fig:network_comparsion}.

GMN achieves the highest F1-score regardless of whether the AST is enriched or not, outperforming all other GNNs. This suggests that the cross-attention mechanism in GMN's propagation effectively enhances the model's ability to capture and compare structural similarities between code snippets. Compared to other GNNs that rely on additional semantic information for improvement, GMN's superior performance indicates that capturing cross-code similarities is more critical for code clone detection than merely enhancing semantic information.

Although GCN and GAT initially perform worse than GGNN with the standard AST representation, their performance significantly improves when enriched with additional semantic information, ultimately reaching a comparable level. This suggests that convolutional and attention-based architectures effectively utilize semantic edge information to enhance structural learning and detection accuracy.

\begin{tcolorbox}
\small
\centering{\bfseries{Key Findings of RQ1}}
\begin{itemize}
    \item \textbf{GMN is the most effective model} due to its cross-attention mechanism, which enhances structural similarity detection. GMN performs well even with the standard AST delivering the highest Precision and nearly the best Recall and F1-score across the board--thus reducing the necessity for hybrid representations. 
    \item \textbf{Hybrid representation effectiveness varies by GNN architecture:} GCN and GAT benefit the most from hybrid representations, while GMN and GGNN show minimal or negative effects. GCN and GAT improve significantly with hybrid representations, reaching performance levels comparable to GGNN.
    \item \textbf{CFG and DFG improve performance, but FA often degrades it.} FA tends to introduce noise in most cases, leading to performance drops--especially in GGNN.
    \item \textbf{GCN and GAT are best suited for hybrid representations,} whereas GMN excels with standard AST, and GGNN struggles with complex structures like FA-AST.
\end{itemize}
\end{tcolorbox}

\subsection{RQ2: What is the computational overhead of different AST-based hybrid representations?}

In this section, we analyze the computational and storage efficiency of various AST-based hybrid graph representations to quantify the overhead introduced by AST enrichment. Specifically, for each representation, we evaluate the generation cost, the storage cost, the graph density, and the inference time when using them with a GNN (in our case, GMN) to assess the trade-offs associated with incorporating additional structural information into AST-based models. These results are shown in Table~\ref{tab:graph_performance_Computational Overhead}.

Based on Table~\ref{tab:graph_performance_Computational Overhead}, we observe significant variations in computational and storage overhead across different AST-based hybrid graph representations. For generation cost, CFG introduces a negligible increase, with AST + CFG requiring 14.224s. However, incorporating DFG results in a large generation cost increase of 21 times, reaching 305.095s. This increase is likely due to the additional complexity involved in tracking data dependencies compared with control dependencies. AST + FA alone requires only 16.910 seconds, reflecting a minor increase compared to AST due to the simple logic to generate it.

The relationship between graph storage cost, average graph density, and inference time is evident in AST-based hybrid representations, as the enriched structural information is primarily encoded through additional edges in the ASTs. As the number of edges increases, the graph storage cost, average graph density, and inference time also increase accordingly. Representations incorporating CFG or DFG introduce a moderate increase in graph density, with AST + CFG and AST + DFG exhibiting densities of 0.0085 and 0.0073, respectively. However, AST + FA representations demonstrate significantly higher graph densities and storage requirements, with AST + FA requiring 1152.00 MB of storage and reaching an average graph density of 0.0202. This increase is due to the inclusion of additional edges representing child-parent relationships, sibling order, token sequences, variable usage, and basic control flow structures such as conditional statements and loops. While these additional structural connections do not explicitly enhance the expressiveness of FA-AST-based representations, their impact on computational performance should be carefully evaluated. 

The increased number of edges in AST-based hybrid representations leads to higher storage requirements, greater graph density, and longer generation and inference times. However, this increase in structural complexity does not necessarily result in improved effectiveness for code clone detection. Therefore, within the research community, it is essential to carefully select the most effective representations when enriching ASTs, ensuring that the additional structural or semantics connections provide meaningful insights while maintaining computational efficiency.

\begin{tcolorbox}
\small
\centering{\bfseries{Key Findings of RQ2}}
\begin{itemize}
    \item \textbf{Generation cost is significantly impacted by DFG, while CFG has a minimal effect.} The generation of AST + CFG requires only $14.224s$, whereas the generation cost of AST + DFG increases to 305.095s, highlighting the high computational cost required for the tracking of data dependencies.
    
    \item \textbf{Structural complexity does not always enhance effectiveness.} While FA enriches representations the most, its computational overhead should be carefully considered for code clone detection tasks.
    \end{itemize}
\end{tcolorbox}

\section{Threats to Validity}
\label{sec:Threats to Validity}
One major limitation of this study is its reliance on the BigCloneBench dataset. While we initially considered using the Google Code Jam (GCJ) \cite{googlecodejam} dataset for comparison, the model's performance on GCJ was exceptionally high, making it unsuitable for a meaningful evaluation. To the best of our knowledge, BigCloneBench remains the most widely used benchmark for code clone detection. However, its focus on Java limits the generalizability of our findings to other programming languages. Mutation-based techniques applied to diverse datasets could potentially enhance generalizability, but this remains an area for future exploration. Additionally, due to the constraints of our experimental setup, cross-language code clone detection was not included in this study.

Another limitation of this study is the scope of evaluation. While AST is the dominant representation in code clone detection, there is a lack of research on the performance of various network architectures applied to different AST designs. This study primarily focuses on AST-based hybrid graphs in the context of GNNs. However, ASTs are also widely used in other architectures, including RNN-based models (RvNN, LSTM, GRU) \cite{fokam2021influence, buch2019learning}, CNNs \cite{patel2021combining}, transformers \cite{zhang2023efficient, hua2022transformer}, and tree-structured models (Tree-LSTM \cite{liang2021ast}, Tree-CNN \cite{yahya2023clcd}). A more comprehensive empirical study exploring AST representations across diverse neural network architectures would provide valuable insights for the research community.

\section{Discussion}
\label{sec:discussion}

Our findings highlight that while certain hybrid representations can enhance clone detection performance, their effectiveness is highly contingent on the underlying GNN architecture. Our results suggest that model architecture—designs incorporating mechanisms such as graph matching—often plays a more critical role than simply enriching ASTs with additional semantic edges. In other words, architectural enhancements may offer more substantial gains than increasing representational complexity alone.

For researchers, this underscores the importance of considering whether the hand-crafted semantic enhancements—such as control or data flow edges—can actually be leveraged by the chosen model. If the architecture lacks the capacity to utilize this information effectively, such enhancements may not only be unhelpful but could even degrade performance.

From a practical perspective, especially in resource-constrained settings, the additional computational and storage overhead introduced by graph enrichment must be weighed carefully. In contrast to controlled experimental environments, real-world applications may benefit more from lightweight configurations.

\section{Conclusion}
\label{sec:conclusion}
In this study, we conducted a systematic evaluation of AST-based hybrid graph representations in the context of GNN-based code clone detection. Our empirical analysis explored the impact of integrating semantic edge information with ASTs, assessing the performance of various GNN architectures, including GCN, GAT, GGNN and GMN.

Our findings reveal that different hybrid representations influence GNN performance in varying ways. While the inclusion of CFGs and DFGs generally improves detection accuracy for models like GCN and GAT, the addition of FAs often introduces complexity that does not necessarily translate to better performance, especially in recurrent architectures like GGNN. Notably, GMN consistently outperformed other models, demonstrating strong capabilities in leveraging AST representations (even in its simplest form, i.e., AST alone) for code clone detection.

This study provides valuable insights for researchers and practitioners aiming to improve code clone detection using GNN-based models. Future work should explore additional hybrid graph structures, extend the evaluation to cross-language code clone detection, and investigate alternative neural architectures such as Transformers. Our open-source implementation serves as a resource for advancing research in this domain.

\section*{Acknowledgment}
This publication has emanated from research conducted with the financial support of Taighde Éireann – Research Ireland under Grant Numbers 18/CRT/6223 and 13/RC/2094\_2.

\bibliographystyle{IEEEtran}
\bibliography{ref}

\begin{thebibliography}{10}
\providecommand{\url}[1]{#1}
\csname url@samestyle\endcsname
\providecommand{\newblock}{\relax}
\providecommand{\bibinfo}[2]{#2}
\providecommand{\BIBentrySTDinterwordspacing}{\spaceskip=0pt\relax}
\providecommand{\BIBentryALTinterwordstretchfactor}{4}
\providecommand{\BIBentryALTinterwordspacing}{\spaceskip=\fontdimen2\font plus
\BIBentryALTinterwordstretchfactor\fontdimen3\font minus \fontdimen4\font\relax}
\providecommand{\BIBforeignlanguage}[2]{{%
\expandafter\ifx\csname l@#1\endcsname\relax
\typeout{** WARNING: IEEEtran.bst: No hyphenation pattern has been}%
\typeout{** loaded for the language `#1'. Using the pattern for}%
\typeout{** the default language instead.}%
\else
\language=\csname l@#1\endcsname
\fi
#2}}
\providecommand{\BIBdecl}{\relax}
\BIBdecl

\bibitem{saini2018code}
N.~Saini, S.~Singh \emph{et~al.}, ``Code clones: Detection and management,'' \emph{Procedia computer science}, vol. 132, pp. 718--727, 2018.

\bibitem{fowler2018refactoring}
M.~Fowler, \emph{Refactoring: improving the design of existing code}.\hskip 1em plus 0.5em minus 0.4em\relax Addison-Wesley Professional, 2018.

\bibitem{kaur2023systematic}
M.~Kaur and D.~Rattan, ``A systematic literature review on the use of machine learning in code clone research,'' \emph{Computer Science Review}, vol.~47, p. 100528, 2023.

\bibitem{liu2023learning}
J.~Liu, J.~Zeng, X.~Wang, and Z.~Liang, ``Learning graph-based code representations for source-level functional similarity detection,'' in \emph{2023 IEEE/ACM 45th International Conference on Software Engineering (ICSE)}.\hskip 1em plus 0.5em minus 0.4em\relax IEEE, 2023, pp. 345--357.

\bibitem{wang2020detecting}
W.~Wang, G.~Li, B.~Ma, X.~Xia, and Z.~Jin, ``Detecting code clones with graph neural network and flow-augmented abstract syntax tree,'' in \emph{2020 IEEE 27th International Conference on Software Analysis, Evolution and Reengineering (SANER)}.\hskip 1em plus 0.5em minus 0.4em\relax IEEE, 2020, pp. 261--271.

\bibitem{fang2020functional}
C.~Fang, Z.~Liu, Y.~Shi, J.~Huang, and Q.~Shi, ``Functional code clone detection with syntax and semantics fusion learning,'' in \emph{Proceedings of the 29th ACM SIGSOFT international symposium on software testing and analysis}, 2020, pp. 516--527.

\bibitem{lu2021code}
Z.~Lu, R.~Li, H.~Hu, and W.-a. Zhou, ``A code clone detection algorithm based on graph convolution network with ast tree edge,'' in \emph{2021 IEEE 21st International Conference on Software Quality, Reliability and Security Companion (QRS-C)}.\hskip 1em plus 0.5em minus 0.4em\relax IEEE, 2021, pp. 1027--1032.

\bibitem{zhao2022precise}
Z.~Zhao, B.~Yang, G.~Li, H.~Liu, and Z.~Jin, ``Precise learning of source code contextual semantics via hierarchical dependence structure and graph attention networks,'' \emph{Journal of Systems and Software}, vol. 184, p. 111108, 2022.

\bibitem{xu2021sccd}
K.~Xu and Y.~Liu, ``Sccd-gan: An enhanced semantic code clone detection model using gan,'' in \emph{2021 IEEE 4th International Conference on Electronics and Communication Engineering (ICECE)}.\hskip 1em plus 0.5em minus 0.4em\relax IEEE, 2021, pp. 16--22.

\bibitem{mehrotra2023improving}
N.~Mehrotra, A.~Sharma, A.~Jindal, and R.~Purandare, ``Improving cross-language code clone detection via code representation learning and graph neural networks,'' \emph{IEEE Transactions on Software Engineering}, 2023.

\bibitem{swilam2023cross}
Z.~Swilam, A.~Hamdy, and A.~Pester, ``Cross-language code clone detection using abstract syntax tree and graph neural network,'' in \emph{2023 International Conference on Computer and Applications (ICCA)}.\hskip 1em plus 0.5em minus 0.4em\relax IEEE, 2023, pp. 1--5.

\bibitem{choi2023investigating}
E.~Choi, N.~Fuke, Y.~Fujiwara, N.~Yoshida, and K.~Inoue, ``Investigating the generalizability of deep learning-based clone detectors,'' in \emph{2023 IEEE/ACM 31st International Conference on Program Comprehension (ICPC)}.\hskip 1em plus 0.5em minus 0.4em\relax IEEE, 2023, pp. 181--185.

\bibitem{kipf2017semi}
T.~N. Kipf and M.~Welling, ``Semi-supervised classification with graph convolutional networks,'' in \emph{International Conference on Learning Representations (ICLR)}, 2017.

\bibitem{velivckovic2018graph}
P.~Veli{\v{c}}kovi{\'c}, G.~Cucurull, A.~Casanova, A.~Romero, P.~Li{\`o}, and Y.~Bengio, ``Graph attention networks,'' in \emph{International Conference on Learning Representations}, 2018.

\bibitem{li2015gated}
Y.~Li, D.~Tarlow, M.~Brockschmidt, and R.~Zemel, ``Gated graph sequence neural networks,'' \emph{arXiv preprint arXiv:1511.05493}, 2015.

\bibitem{li2016gated}
------, ``Gated graph sequence neural networks,'' in \emph{Proceedings of the International Conference on Learning Representations (ICLR)}, 2016.

\bibitem{yuan2022java}
D.~Yuan, S.~Fang, T.~Zhang, Z.~Xu, and X.~Luo, ``Java code clone detection by exploiting semantic and syntax information from intermediate code-based graph,'' \emph{IEEE Transactions on Reliability}, vol.~72, no.~2, pp. 511--526, 2022.

\bibitem{svajlenko2015evaluating}
J.~Svajlenko and C.~K. Roy, ``Evaluating clone detection tools with bigclonebench,'' in \emph{2015 IEEE international conference on software maintenance and evolution (ICSME)}.\hskip 1em plus 0.5em minus 0.4em\relax IEEE, 2015, pp. 131--140.

\bibitem{svajlenko2016bigcloneeval}
------, ``Bigcloneeval: A clone detection tool evaluation framework with bigclonebench,'' in \emph{2016 IEEE international conference on software maintenance and evolution (ICSME)}.\hskip 1em plus 0.5em minus 0.4em\relax IEEE, 2016, pp. 596--600.

\bibitem{van2020clone}
B.~van Bladel and S.~Demeyer, ``Clone detection in test code: an empirical evaluation,'' in \emph{2020 IEEE 27th International Conference on Software Analysis, Evolution and Reengineering (SANER)}.\hskip 1em plus 0.5em minus 0.4em\relax IEEE, 2020, pp. 492--500.

\bibitem{wang2023comparison}
Y.~Wang, Y.~Ye, Y.~Wu, W.~Zhang, Y.~Xue, and Y.~Liu, ``Comparison and evaluation of clone detection techniques with different code representations,'' in \emph{2023 IEEE/ACM 45th International Conference on Software Engineering (ICSE)}.\hskip 1em plus 0.5em minus 0.4em\relax IEEE, 2023, pp. 332--344.

\bibitem{zubkov2022evaluation}
M.~Zubkov, E.~Spirin, E.~Bogomolov, and T.~Bryksin, ``Evaluation of contrastive learning with various code representations for code clone detection,'' \emph{arXiv preprint arXiv:2206.08726}, 2022.

\bibitem{zhang2023efficient}
A.~Zhang, L.~Fang, C.~Ge, P.~Li, and Z.~Liu, ``Efficient transformer with code token learner for code clone detection,'' \emph{Journal of Systems and Software}, vol. 197, p. 111557, 2023.

\bibitem{svajlenko2014towards}
J.~Svajlenko, J.~F. Islam, I.~Keivanloo, C.~K. Roy, and M.~M. Mia, ``Towards a big data curated benchmark of inter-project code clones,'' in \emph{2014 IEEE International Conference on Software Maintenance and Evolution}.\hskip 1em plus 0.5em minus 0.4em\relax IEEE, 2014, pp. 476--480.

\bibitem{wei2017supervised}
H.~Wei and M.~Li, ``Supervised deep features for software functional clone detection by exploiting lexical and syntactical information in source code.'' in \emph{IJCAI}, 2017, pp. 3034--3040.

\bibitem{paszke2019pytorch}
A.~Paszke, S.~Gross, F.~Massa, A.~Lerer, J.~Bradbury, G.~Chanan, T.~Killeen, Z.~Lin, N.~Gimelshein, L.~Antiga \emph{et~al.}, ``Pytorch: An imperative style, high-performance deep learning library,'' \emph{Advances in neural information processing systems}, vol.~32, 2019.

\bibitem{fey2019fast}
M.~Fey and J.~E. Lenssen, ``Fast graph representation learning with pytorch geometric,'' \emph{arXiv preprint arXiv:1903.02428}, 2019.

\bibitem{googlecodejam}
2016, google Code Jam https://code.google.com/codejam/contests.html.

\bibitem{fokam2021influence}
M.~A. Fokam and R.~Ajoodha, ``Influence of contrastive learning on source code plagiarism detection through recursive neural networks,'' in \emph{2021 3rd International Multidisciplinary Information Technology and Engineering Conference (IMITEC)}.\hskip 1em plus 0.5em minus 0.4em\relax IEEE, 2021, pp. 1--6.

\bibitem{buch2019learning}
L.~B{\"u}ch and A.~Andrzejak, ``Learning-based recursive aggregation of abstract syntax trees for code clone detection,'' in \emph{2019 IEEE 26th International Conference on Software Analysis, Evolution and Reengineering (SANER)}.\hskip 1em plus 0.5em minus 0.4em\relax IEEE, 2019, pp. 95--104.

\bibitem{patel2021combining}
S.~Patel and R.~Sinha, ``Combining holistic source code representation with siamese neural networks for detecting code clones,'' in \emph{IFIP International Conference on Testing Software and Systems}.\hskip 1em plus 0.5em minus 0.4em\relax Springer, 2021, pp. 148--159.

\bibitem{hua2022transformer}
W.~Hua and G.~Liu, ``Transformer-based networks over tree structures for code classification,'' \emph{Applied Intelligence}, vol.~52, no.~8, pp. 8895--8909, 2022.

\bibitem{liang2021ast}
H.~Liang and L.~Ai, ``Ast-path based compare-aggregate network for code clone detection,'' in \emph{2021 International Joint Conference on Neural Networks (IJCNN)}.\hskip 1em plus 0.5em minus 0.4em\relax IEEE, 2021, pp. 1--8.

\bibitem{yahya2023clcd}
M.~A. Yahya and D.-K. Kim, ``Clcd-i: cross-language clone detection by using deep learning with infercode,'' \emph{Computers}, vol.~12, no.~1, p.~12, 2023.

\end{thebibliography}

\end{document}